\documentclass[11pt]{article}

\usepackage[final]{acl}

\usepackage{times}
\usepackage{latexsym}

\usepackage[T1]{fontenc}

\usepackage[utf8]{inputenc}

\usepackage{microtype}

\usepackage{inconsolata}

\usepackage{graphicx}
\usepackage{subcaption}
\usepackage{tabularx} 
\usepackage{multirow} 
\usepackage[toc,page]{appendix}
\usepackage{algorithm}
\usepackage{algpseudocode}
\usepackage{amsmath}
\algnewcommand{\Input}{\State \textbf{Input: }}
\algnewcommand{\Output}{\State \textbf{Output: }}

\usepackage{booktabs}
\usepackage{enumitem}
\usepackage{kotex}
\usepackage{amsmath}
\usepackage{graphicx}
\usepackage{tcolorbox}
\usepackage{makecell}
\usepackage{float}

\newcommand{\sbullet}[1][.8]{\mathbin{\vcenter{\hbox{\scalebox{#1}{$\bullet$}}}}}

\title{Leveraging External Knowledge for Historical Document Restoration via Retrieval-Augmented Large Language Models}

\author{
 \textbf{Gabeen Kim\textsuperscript{1}} \qquad
 \textbf{Kyeongpil Kang\textsuperscript{2}\thanks{Corresponding author}}
\\
\\
 \textsuperscript{1}Department of AI Convergence, Kangwon National University \\
 \textsuperscript{2}Department of Computer Science and Engineering, Kangwon National University
\\
 \normalsize\texttt{gokong0516@kangwon.ac.kr \quad rudvlf0413@kangwon.ac.kr}
}

\begin{document}
\maketitle

\begin{abstract}
Historical documents act as invaluable knowledge archives but often suffer from illegibility due to physical deterioration and damage. While existing restoration methods based on masked language modeling effectively utilize local context, they struggle to restore named entities that require external historical knowledge. To address this limitation, we introduce a novel framework for historical document restoration that leverages large language models with retrieval-augmented generation (RAG). By combining the implicit knowledge of pre-trained LLMs with explicitly retrieved external context, our model ARI effectively mitigates the challenge of inferring context-dependent proper nouns. Extensive experiments on Korean historical documents demonstrate that our approach significantly outperforms baselines, achieving substantial gains in restoring both general characters and named entities. Furthermore, comprehensive evaluations including expert assessments confirm that \textbf{ARI} serves as a practical tool for domain experts, promising to accelerate the analysis of historical records.
\end{abstract}
\section{Introduction}
\label{sec:intro}
Historical records are repositories of vast information spanning centuries or even millennia. Recognizing the global importance of these records, national governments and researchers are working not only to preserve these ancient documents but also to uncover and analyze the knowledge they contain. For instance, the Annals of the Joseon Dynasty (\textbf{AJD})\footnote{\url{https://sillok.history.go.kr/intro/english.do}} and the Journal of the Royal Secretariat (\textbf{JRS})\footnote{\url{https://sjw.history.go.kr/intro/engInfo.do}}, which span 500 years of history and contain not only historical facts but also natural events, are inscribed on the UNESCO Memory of the World Register. These archives are extensively analyzed by researchers across diverse domains to derive valuable insights.

\begin{figure}[t]
    \centering
    \includegraphics[width=\columnwidth]{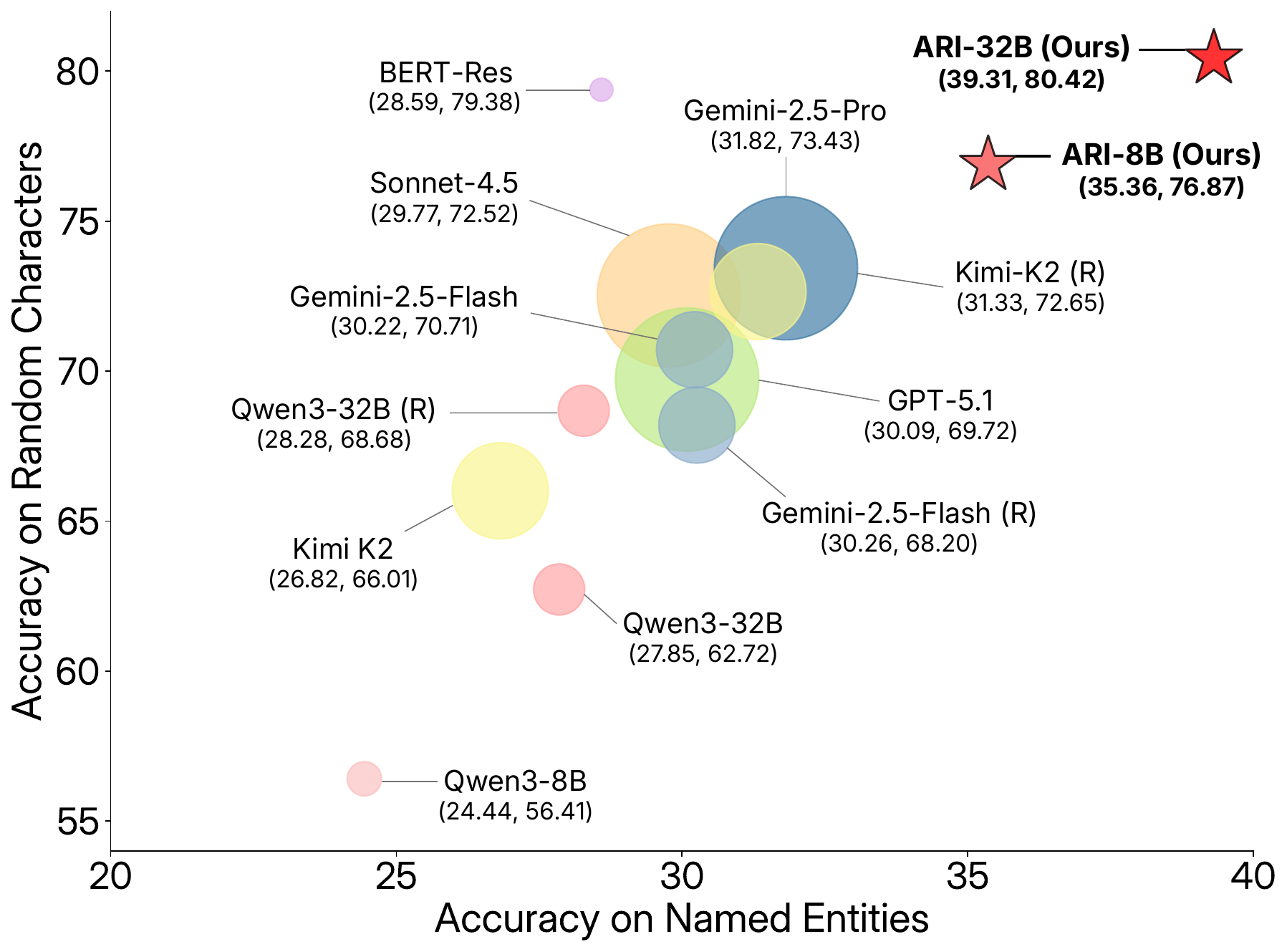}
    \caption{Performance of the proposed model compared with the baselines. The x- and y-axes represent the restoration accuracies for named entities and randomly masked characters, respectively.}
    \label{fig:overall_performance}
\end{figure}

While these historical archives possess immense value, preserving them and extracting knowledge present significant challenges. A primary issue is the inherent vulnerability of historical manuscripts to damage and degradation. These documents were typically inscribed on materials such as wood or plant fibers, which are far less durable than modern paper. Consequently, they are highly sensitive to environmental conditions, prone to discoloration and structural deterioration upon exposure to light, heat, or moisture. Furthermore, prior to the advent of metal movable type, texts were transcribed manually using brushes or pens. Accordingly, legibility was highly contingent on the scribe's penmanship, which inevitably introduced transcription errors. Such compromised sections are difficult for both OCR systems and human experts to analyze. As a result, they are often treated as blanks or damaged tokens, which significantly hinders downstream analysis. For instance, by matching those damaged tokens, we identified that approximately 41.9K characters across 11.1K documents in the JRS remain damaged or unrecognizable.

To restore illegible parts of historical archives, several approaches have been proposed. \citet{kang2021restoring} addressed the restoration of damaged documents by training a model via masked language modeling (MLM). However, a key limitation of these methods is their exclusive reliance on the damaged document itself, neglecting external information. Integrating external knowledge is crucial for correctly inferring named entities, particularly proper nouns. For example, in the sentence ``In 1492, Columbus first landed in \textbf{[M]}.'', predicting the masked token \textbf{[M]} is challenging using only the local context. In such cases, the model must either possess internalized historical knowledge or utilize external resources to ensure accurate restoration.

To overcome this limitation, we introduce a novel framework for historical document restoration utilizing LLM-based retrieval-augmented generation. Our method utilizes a pre-trained LLM that has indirectly learned from extensive datasets including historical knowledge and documents available on the web. This implicitly learned context acts as vital external knowledge, significantly improving the restoration of damaged characters. Additionally, we integrate RAG to allow the model to explicitly access external knowledge by incorporating relevant documents into the prompt. Finally, we developed our model, \textbf{ARI} (\textbf{A}rchive \textbf{R}estoration \textbf{I}ntelligence), specifically trained on ancient corpora. Experimental results demonstrate that our model outperforms baselines. We have made our models and code publicly available to facilitate future research.\footnote{\url{https://github.com/EvelynKimm/ARI}}

\begin{figure}[t]
    \centering
    \includegraphics[width=\columnwidth]{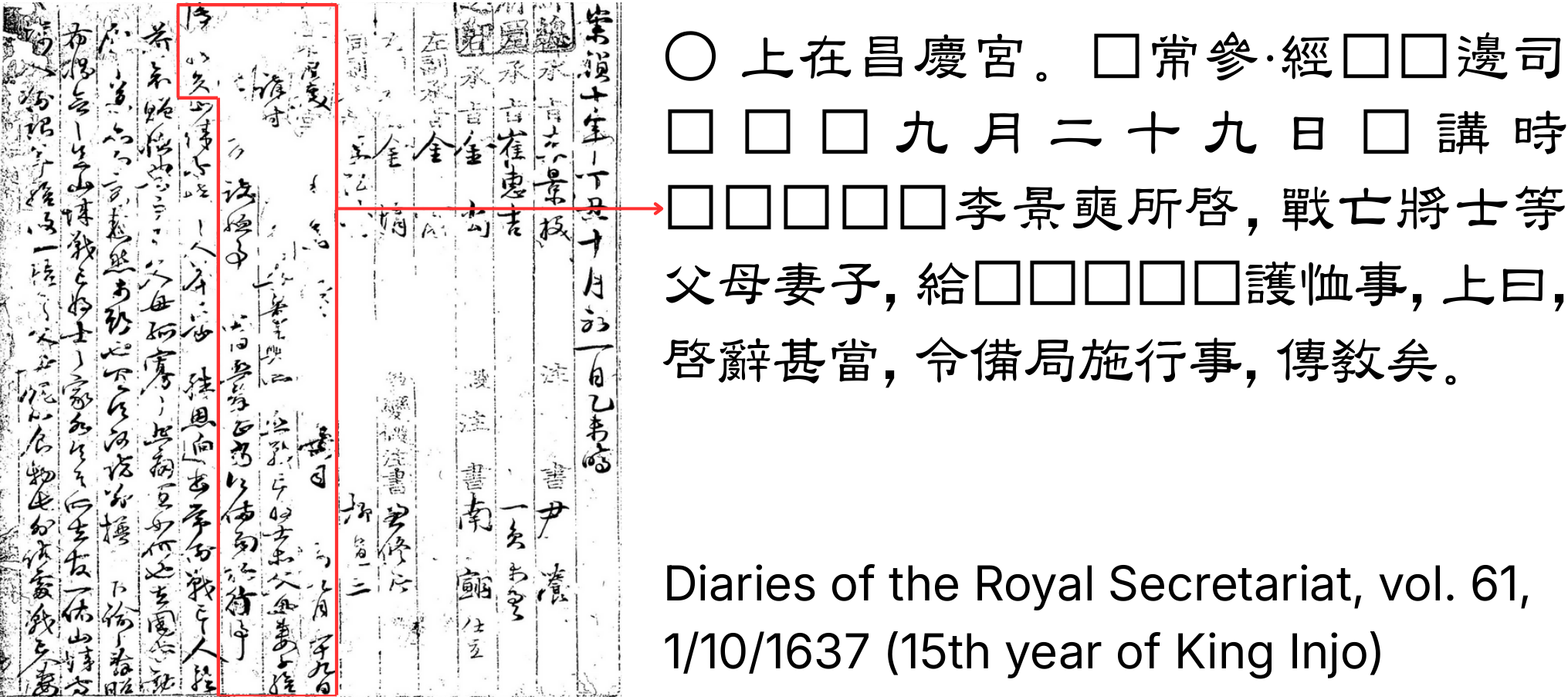}
    \caption{Real-world example of a damaged historical document with masked characters (□).}
    \label{fig:damaged_example}
\end{figure}

\section{Related Work}
\label{sec:related_work}

Ancient documents serve as crucial data sources with high research value across diverse domains, ranging from history to the social and natural sciences. However, as previously discussed, many historical records have been damaged  from various factors, inevitably limiting their research utility.

Accordingly, methods have been proposed to restore these damaged parts. Initial efforts focused on deep learning approaches to correct typographical mistakes in historical texts~\cite{tang2018evaluation,domingo2018spelling}. Pythia, designed to predict missing characters in Greek epigraphy, achieved a top-20 accuracy of 73.5\%~\cite{assael2019restoring}. \citet{kang2021restoring} achieved both effective restoration and translation by jointly training MLM and machine translation. While Ithaca enhanced interpretability by jointly predicting textual, geographical, and chronological attributes~\cite{assael2022restoring}, our target corpora (AJD and JRS) inherently lack geographical metadata. Leveraging the improved general performance and linguistic capabilities of pretrained LLMs, \citet{liu2025zero} demonstrated that restoration is feasible in a zero-shot fashion.

However, prior works have restricted the model input to a single corrupted document, limiting predictions to the document's internal context. As a result, these methods exhibit significant shortcomings in restoring information that necessitates external knowledge, such as named entities. This highlights the necessity of leveraging external data sources to effectively restore damaged historical documents. Another limitation stems from the fixed masking probability used during training~\cite{kang2021restoring,assael2022restoring}. This creates a discrepancy, as the masking rate does not match the corruption distribution in actual documents. This mismatch between training and inference conditions can degrade the real-world performance.

Language models trained on extensive web data demonstrate superior general performance. BERT~\cite{devlin2019bert}, for example, utilizes a bi-directional Transformer Encoder trained with MLM that predicts original tokens by masking random positions in the training data. Given its resemblance to recovering damaged text, MLM has been extensively used for historical document restoration. However, BERT-based approaches have a key drawback: They predict masked tokens in a non-autoregressive manner, often missing the dependencies among them. Recently, causal language modeling has become the standard pre-training method for the latest LLMs, such as Gemini-2.5~\cite{comanici2025gemini}, Sonnet 4.5~\cite{anthropic2025claude-sonnet-4-5}, Kimi-K2~\cite{team2025kimi}, GPT-5.1~\cite{openai_gpt5_1}, and Qwen3~\cite{yang2025qwen3}, due to the simplicity of its training objective. Effectively ingesting extensive linguistic data and general knowledge, these models internalize implicit knowledge from vast archives (e.g., Chinese documents, Korean Hanja texts, and historical records). This model-intrinsic knowledge can then be leveraged as a powerful source of external knowledge for the restoration of damaged historical documents.

Additionally, RAG has been developed to improve LLM performance by directly integrating external knowledge. RAG operates by retrieving documents relevant to a user's query via lexical-based (e.g., BM25) or embedding-based methods from external repositories like the web or databases~\cite{lewis2020retrieval}. Providing these documents as context enables the model to utilize information not learned during pre-training and effectively alleviates hallucinations. RAG is particularly effective for domain-specific and knowledge-intensive tasks, such as those involving low-resource languages~\cite{nie2023cross,chang2025enhancing} and named entity recognition~\cite{xie2025retrieval}. In addition, research suggests that applying fine-tuning to RAG systems enables them to better leverage relevant documents, resulting in significant performance gains on downstream tasks~\cite{zhangraft}.

In this study, we propose a method for effectively restoring historical documents by leveraging external knowledge. We utilize LLM that employs a causal language modeling approach to model the contextual dependencies of damaged characters. Additionally, we employ RAG to provide relevant documents to the model, thereby allowing it to draw on external knowledge. This approach is further refined through fine-tuning, which optimizes the model's capacity to leverage external information and enhances final restoration accuracy.
\section{Dataset}
\label{sec:dataset}

\begin{table}[t]
\centering
\begin{tabular}{lcc}
\toprule
                         & AJD & JRS \\
\midrule
Number of Docs.             & 0.37M & 1.75M \\
Number of Chars.      & 71.9M & 292.6M \\
Avg. NEs per Doc. & 1.79 & 4.73 \\
\midrule
Number of Docs. w/ [D] & 56 & 11.1K \\
Number of [D]                & 0.11K & 41.9K \\
Avg. [D] per Doc.   & 1.98 & 3.77 \\
\bottomrule
\end{tabular}

\caption{Statistics of the AJD and JRS raw data corpora. NEs and [D] represent named entities and damaged tokens, respectively.}
\label{tab:raw_data_statistics}
\end{table}

\begin{figure}[t]
    \centering
    \includegraphics[width=\columnwidth]{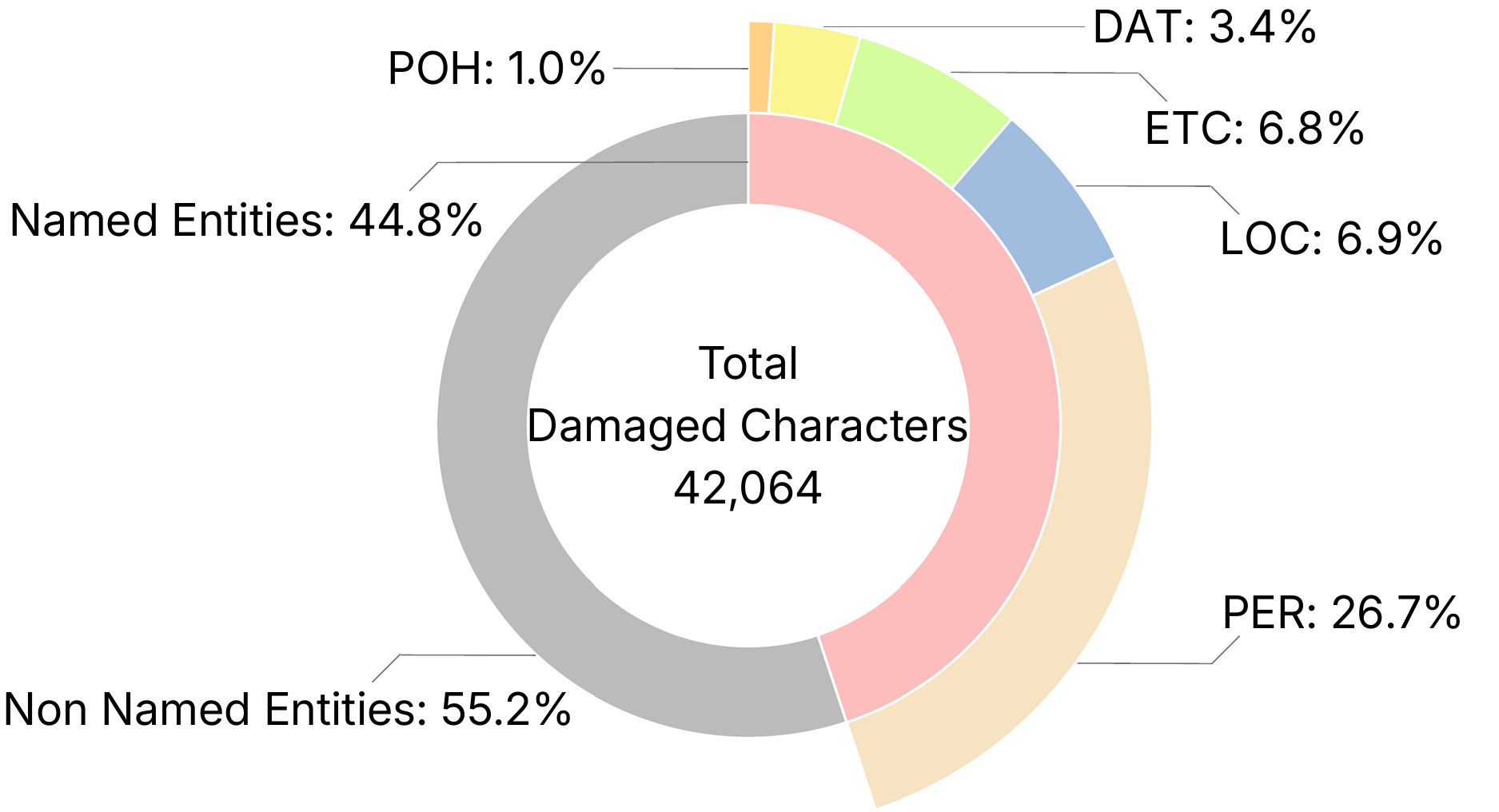}
    \caption{Pie chart depicting the distribution of damaged characters, including the categorical ratios of named entities: PER, LOC, DAT, POH, and ETC.}
    \label{fig:named_entity_distribution}
\end{figure}

In this study, we utilized AJD and JRS to train and evaluate our restoration model. Both corpora consist of documents written in Hanja and contain rich metadata, including temporal information (e.g., year, month, and date) and named entities (e.g., personal names, organizations, book titles) provided by the National Institute of Korean History.\footnote{\url{https://www.history.go.kr/en/main/main.do}} Table~\ref{tab:raw_data_statistics} presents the statistics for the raw data. We observed a significantly higher prevalence of damaged characters in the JRS compared to the AJD. Furthermore, within the documents containing damage, the average number of damaged characters was 3.77 per document. To investigate the characteristics of these damaged characters in detail, we employed Gemini-2.5-Pro to predict the named entity type for each character.\footnote{To evaluate the reliability of the entity classification labels generated by Gemini-2.5-Pro, we manually inspected the entities in 200 randomly sampled documents. The results showed that only 4 instances (0.7\%) were misclassified.} As illustrated in Fig. \ref{fig:named_entity_distribution}, approximately 44.8\% of the 42K damaged characters were identified as named entities. A breakdown of the categories reveals that PER (Person) was the most frequent, followed by LOC (Location), ETC (Organizations, Official Titles, etc.), DAT (Date), and POH (Publication of History). While Gemini-2.5-Pro may not be perfectly accurate, these results function as a proxy, indirectly reflecting the distribution trends regarding the presence and types of named entities. Therefore, these findings underscore the importance of character restoration not merely for arbitrary positions, but specifically for named entity segments.

We strictly held out documents containing real-world corrupted parts for the human-evaluation dataset $D_{\text{RD}}$. The remaining data comprised 10K documents each for testing and validation, with the other 2.02M used for training. To prevent data leakage, we strictly removed from the training corpus any samples duplicated in the validation or test sets. This training corpus acted as the retrieval source for the RAG system and was used to train our model and baselines. For building the test dataset using the test corpus, we introduced synthetic corruptions at random positions, taking into account the distribution of corruption span lengths observed in real damaged documents. We set the corruption rate stochastically to best emulate the original data's corruption distribution. Specifically, the rate was centered around a mean of 2.96\% (the average rate from the original corrupted documents), and we guaranteed a minimum of one corrupted character per document. In addition, assessing the reconstruction accuracy of the damaged parts is crucial, especially in cases requiring external data. Accordingly, we designed our valid and test datasets to comprise the following two versions:
\begin{itemize}
    \item $D_{\text{Rand}}$: Dataset where synthetic corruptions are applied at random positions. This dataset serves to assess the overall restoration performance of the model.
    \item $D_{\text{NE}}$: Dataset containing synthetic corruptions originating from named entities. This dataset is used to evaluate the model's ability to restore segments requiring external knowledge.
\end{itemize}

These evaluation datasets consist of pairs of partially corrupted input documents and their corresponding ground-truth originals. The data construction pipeline is detailed in Section~\ref{sec:proposed_methods}.
\section{Proposed Methods}
\label{sec:proposed_methods}

\begin{figure*}[t]
    \centering
    \includegraphics[width=\linewidth]{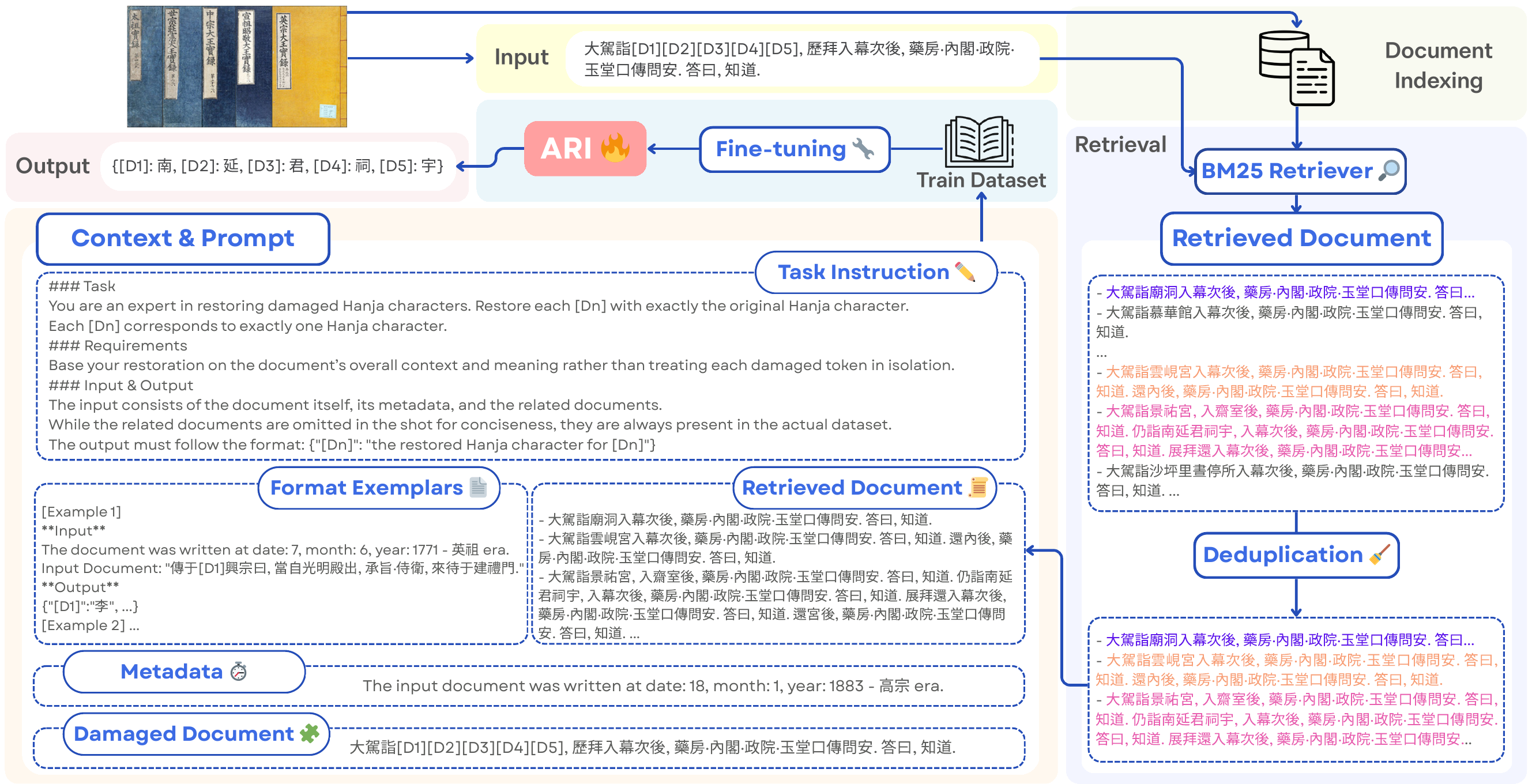}
    \caption{Overview of the proposed framework for restoring damaged documents. The inputs to the LLM include a task description, examples of formats, documents related to the damaged text, and metadata such as temporal information. Based on this structure, ARI is fine-tuned specifically for the historical document restoration task.}
    \label{fig:proposed_method}
\end{figure*}

As shown in Fig.~\ref{fig:proposed_method}, this section presents our restoration framework. To validate the framework, we first evaluate the performance of baseline models. We then demonstrate how restoration accuracy can be improved by leveraging external knowledge via metadata and RAG. Finally, we present our proposed model, ARI, alongside the baseline, BERT-Res trained from scratch on our dataset.

\subsection{LLM Baseline Restoration Performance} \label{llm_baseline}

\begin{figure}
    \centering
    {\fontsize{10pt}{12.pt}\selectfont
    \begin{tcolorbox}[
        width=1.0\linewidth,   
        colback=gray!5,         
        colframe=gray!60!black, 
        boxrule=1pt,          
        arc=1mm,                
        boxsep=1pt,             
        left=4pt, right=4pt,    
        top=5pt, bottom=5pt,    
        before skip=4pt,        
        after skip=12pt          
    ]
    \ttfamily
    \raggedright    
    \textbf{System prompt:} Restore [Dn] with the original Hanja character. Each [Dn] corresponds to exactly one Hanja character. Output must be a valid json object with the following format:
    
    \{``[Dn]'': ``Restored character for [Dn]''\}
    \vspace{2mm}
    
    \textbf{Input:} "A[D1]CDE[D2]G"
    
    \vspace{1mm}
    
    \textbf{Output:} \texttt{\{"[D1]": "B", "[D2]": "F"\}}
    \end{tcolorbox}
    }
    \caption{An example of the base prompt for LLM-based restoration.}
    \label{fig:base_prompt}
\end{figure}

\citet{kang2021restoring} utilized a Transformer Encoder model to directly restore original characters at each masked position. However, given the autoregressive nature of most pre-trained language models, it is crucial to explicitly identify the index (position) of each corrupted character and its corresponding restoration within the input and output. Therefore, as shown in Fig.~\ref{fig:base_prompt}, we designed the base prompt format to incorporate positional information for each corrupted character and extract the restored characters accordingly.

We then evaluated the basic restoration performance by measuring character prediction accuracy on the valid dataset across multiple LLMs. The accuracy for each model is presented in Table~\ref{tab:llm_base_performance}. Within the Qwen3 and Gemini-2.5 families, we observed a positive correlation between model scale and restoration accuracy. Notably, the results from Kimi K2 and Gemini-2.5-Flash demonstrate that enabling thinking mode fosters deeper reasoning, thereby enhancing named entity restoration capabilities. However, a slight decline was observed in the restoration performance for random characters. Gemini-2.5-Pro and Sonnet 4.5 surpassed other models in restoring characters at random positions, reflecting their linguistic and contextual understanding of Hanja characters. For named entities, Sonnet 4.5 achieved the highest performance. This suggests that Sonnet 4.5 likely engaged in more extensive learning of historical background knowledge during the pre-training phase compared to other models.

\begin{table}
    \centering
    \setlength{\tabcolsep}{5.8pt}
    \begin{tabular}{lccc}
        \toprule
         \makecell[c]{Model} & Thinking & \makecell{Acc.\\(NE)} & \makecell{Acc.\\(Rand)} \\
         \midrule
         Qwen3 8B & No & 2.86 & 7.43 \\
         Qwen3 32B & No & 5.57 & 13.37 \\
         Kimi K2 & No & 5.83 & 22.45 \\
         Kimi K2 & Yes & 6.53 & 20.66 \\
         Gemini-2.5-Flash & No & 6.82 & 24.38 \\
         Gemini-2.5-Flash & Yes & 7.10 & 22.35 \\
         Gemini-2.5-Pro & Yes & 9.57 & \textbf{32.62} \\
         GPT-5.1 & Yes & 10.03 & 28.69 \\
         Sonnet 4.5 & Yes & \textbf{13.30} & 32.35 \\
         \bottomrule
    \end{tabular}
    \caption{Reconstruction performance (Top-1 accuracy) of LLMs on $D_{\text{NE}}$ and $D_{\text{Rand}}$ with the base prompt.}
    \label{tab:llm_base_performance}
\end{table}

\subsection{Enhancing Restoration Performance via External Knowledge} \label{external_knowledge}

\begin{table}[t]
    \centering
    \begin{tabular}{lcc}
        \toprule
          & Acc. of NE & Acc. of Rand \\
         \midrule
         Baseline & 5.57 & 13.37 \\
         + Metadata & \textbf{5.63} & \textbf{14.83} \\
         \bottomrule
    \end{tabular}
    \caption{Restoration performance results based on the inclusion of metadata.}
    \label{tab:performance_metadata}
\end{table}

External knowledge plays a crucial role in the precise restoration of damaged characters. To leverage this, we propose an enhanced prompting strategy that integrates chronological metadata directly into the LLM context. Specifically, we augment the input with temporal identifiers such as the reigning king, year, month, and day. As detailed in Table~\ref{tab:performance_metadata}, this integration improves the model's reconstruction accuracy, validating that temporal grounding is essential for effective restoration.

\begin{table}[t]
    \centering
    \begin{tabular}{lcc}
        \toprule
        & Acc. of NE & Acc. of Rand \\
        \midrule
        Baseline & 5.63 & 14.83 \\
        \midrule
        + Static shots & 6.27 & 18.98 \\
        \midrule
        + Dynamic shots & & \\
        \quad$\sbullet$ Random & 8.55 & 26.18 \\
        \quad$\sbullet$ Embedding & 19.88 & 59.91 \\
        \quad$\sbullet$ BM25 & 24.28 & 60.36 \\
        \midrule
        + Deduplication & \textbf{27.85} & \textbf{61.45} \\
        \bottomrule
    \end{tabular}
    \caption{Impact of adding few-shot, RAG, and deduplication on restoration performance.}
    \label{tab:rag_ablation_study}
\end{table}

Beyond simply adding metadata, a key contribution of our approach is the strategic integration of external knowledge via few-shot prompting and Retrieval-Augmented Generation. We specifically designed our experiments to demonstrate how injecting this external context augments the restoration capabilities of the Qwen3 32B. The significant performance improvements yielded by these techniques are detailed in Table~\ref{tab:rag_ablation_study}.

To ensure the model's strict adherence to the required output schema, we initially incorporated five few-shot format exemplars into the system prompt. These examples comprise distinct input-output pairs following a format similar to that shown in Fig.~\ref{fig:base_prompt}. Empirical results indicate that providing these format guides significantly reduces formatting errors and aligns the model's generation with the intended restoration protocol.

Furthermore, to incorporate external knowledge via RAG, the top 20 most relevant documents for each input were retrieved from the training corpus. We evaluated three retrieval methods: random selection, embedding-based retrieval, and BM25. Specifically, we used Gemini Embedding~\cite{lee2025gemini} for dense retrieval and BM25S~\cite{lu2024bm25s} for sparse retrieval. The results indicate that even randomly selecting reference documents improves performance compared to using no references. These random examples function as few-shot demonstrations, providing the model with general knowledge regarding Hanja and its usage. However, both embedding-based retrieval and BM25 significantly outperformed random selection, confirming the effectiveness of retrieving contextually relevant documents. Notably, BM25 achieved the highest performance among the three methods. This suggests that character matching is particularly effective for restoring damaged Hanja characters, due to the logographic nature of Hanja where each character encapsulates a specific concept. Further analysis of the retrieval strategies is provided in Appendix~\ref{sec:retrieval_analysis}.

\begin{figure}[t]
    \centering
    \includegraphics[width=\columnwidth]{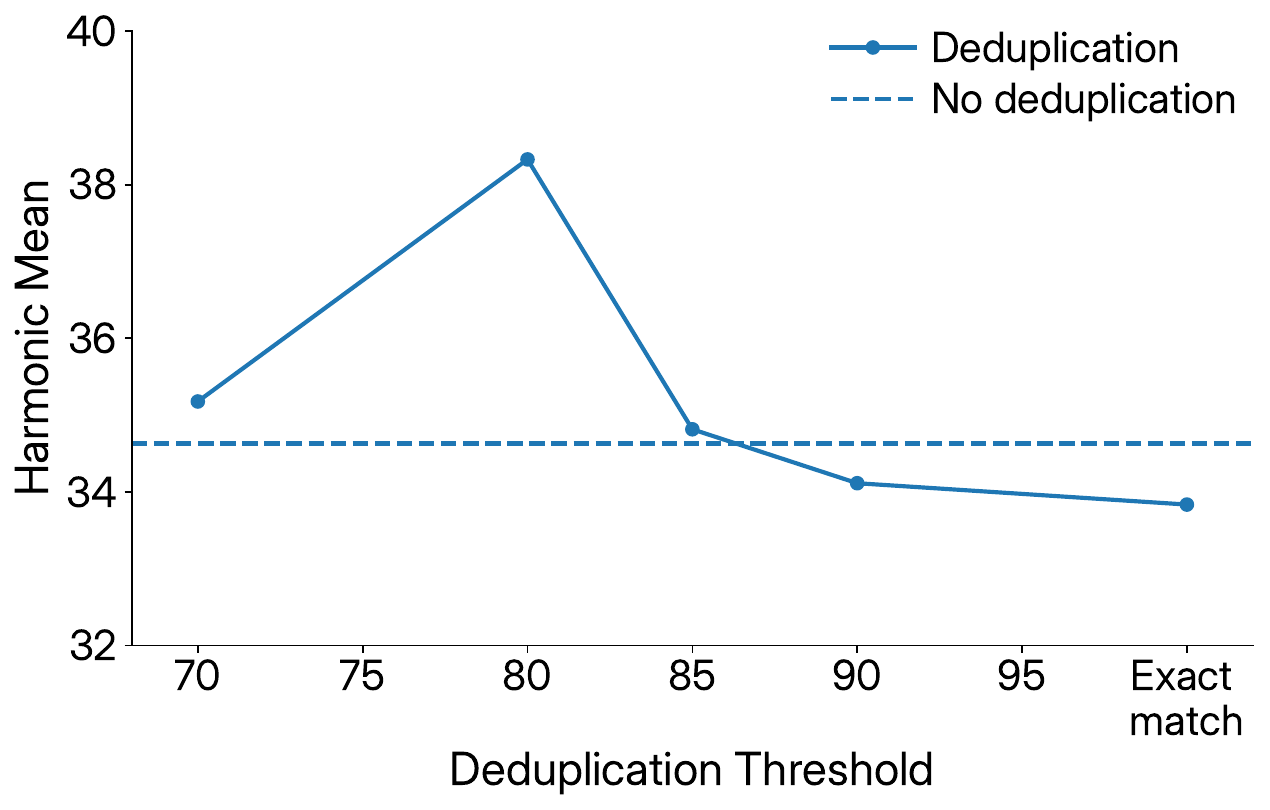}
    \caption{Restoration performance across deduplication thresholds. The plot represents the harmonic mean of $D_{\text{NE}}$ and $D_{\text{Rand}}$ restoration performances.}
    \label{fig:deduplication_threshold}
\end{figure}

Furthermore, the corpus contains duplicates and near-duplicates, which are common in short texts. Retaining such redundant content in the retrieved documents limits the model's access to diverse information and risks introducing bias. Consequently, we employ a deduplication method based on string similarity\footnote{\url{https://github.com/rapidfuzz/RapidFuzz}} among the retrieved documents. Fig.~\ref{fig:deduplication_threshold} illustrates the impact of various deduplication thresholds on restoration performance. The solid line depicts the harmonic mean of the performance on $D_{\text{NE}}$ and $D_{\text{Rand}}$, while the dotted line indicates the baseline without deduplication.

Significant improvements over the baseline are observed at thresholds of 70\% and 80\%, with performance peaking at 80\%. However, performance declines as the threshold exceeds 85\%, reaching its lowest point at the exact-match threshold (100\%), where only identical documents are removed. This suggests that preserving highly similar documents restricts the scope of external knowledge and biases the model toward duplicates, thereby degrading restoration performance. By contrast, ensuring diversity among relevant documents through effective deduplication enhances performance by providing richer contexts and cues for restoration. Therefore, we filter out documents that exceed a similarity threshold of 80\%. This approach yields further performance improvements in restoring both named entities and randomly positioned content.

Overall, our experiments indicate that few-shot prompting and RAG act synergistically to enhance LLM-based restoration. This combination facilitates task comprehension and external knowledge retrieval, which are critical for accurately restoring damaged characters. The detailed prompt format is provided in Fig.~\ref{fig:final_prompt} in Appendix~\ref{sec:prompt}.

\subsection{Fine-Tuning the ARI Model}
To construct the training dataset for fine-tuning, we masked characters at random positions within a training corpus explicitly isolated from the test corpus. As shown in Table~\ref{tab:train_mixing_ratio}, prioritizing named entities masking in 25\% of the training data enhanced restoration performance on $D_{\text{NE}}$ while maintaining performance on $D_{\text{Rand}}$. Accordingly, we applied this named entity-prioritized masking strategy to 25\% of the training dataset.

We then fine-tuned an open-source LLM with the prompt format described above to build a restoration model specialized for the Hanja domain. To mitigate overfitting and improve pattern diversity, we adopted a dynamic masking strategy in which mask positions are varied across epochs, following RoBERTa~\cite{liu2019roberta}. Finally, we developed our model, \textbf{ARI-32B}, by fine-tuning Qwen3 32B, incurring a computational cost of approximately 1,500 H200 GPU hours. In addition, we trained a computationally efficient variant, \textbf{ARI-8B}, based on Qwen3 8B, while maintaining competitive performance against other baseline models.

Additionally, we implemented a baseline following \citet{kang2021restoring} that relies solely on the damaged input text, without access to external context. We denote the model that prioritizes named entities for masking at a 25\% rate as \textbf{BERT-Res}. We employed ModernBERT-large~\cite{warner2025smarter} for its architectural advancements over the original BERT. Since the original ModernBERT vocabulary lacks sufficient coverage for Hanja characters, we extended the tokenizer with Hanja characters. The model was then trained from scratch on our restoration dataset for 10 epochs to ensure convergence. This baseline therefore provides an appropriate setting for evaluating the impact of external knowledge, as it performs restoration without access to external context. Training details, including hyperparameters for our model and the baseline, are provided in Appendix~\ref{sec:training_details}.

\section{Experimental Results}
\label{sec:experimental_results}

\subsection{Evaluation on Test Dataset}

To evaluate our model against the BERT-Res baseline and other LLMs, we measured top-$K$ accuracy on $D_{\text{Rand}}$ and $D_{\text{NE}}$. Specifically, we employed the identical input prompt for both our model and the open-source/proprietary LLMs to ensure the consistent utilization of external knowledge. As shown in Fig.~\ref{fig:overall_performance}, the restoration performance for $D_{\text{NE}}$ is significantly lower than that for $D_{\text{Rand}}$. This implies that restoring named entities is more challenging than restoring general characters.

Notably, LLMs utilizing only external knowledge without fine-tuning outperformed BERT-Res in named entity restoration. This demonstrates that incorporating external knowledge effectively enhances restoration capabilities for named entities. Regarding model scale, the Qwen3 32B exhibited superior performance compared to the 8B variant within the same family, confirming that model size significantly impacts performance. With respect to reasoning strategies, we observed performance improvements across Qwen3 32B, Kimi K2, and Gemini-2.5-Flash in random position restoration tasks. Similar improvements were noted in named entity restoration, with the exception of Gemini-2.5-Flash. These findings suggest that reasoning mechanisms enable LLMs to better leverage knowledge acquired during pre-training, facilitating the resolution of complex restoration tasks.

The BERT-Res model, trained from scratch without external knowledge, outperformed Gemini-2.5-Pro in the general character restoration task. This result underscores the importance of training for restoring general characters in historical documents. Furthermore, despite having fewer parameters, the model surpassed decoder-based models, likely due to its bidirectional encoder architecture. However, its performance on named entities fell short of other LLMs, suggesting that relying solely on intra-document context is insufficient. Ultimately, the highest restoration performance was achieved by ARI-32B, which integrates external knowledge via RAG with fine-tuning. Our model demonstrates superior performance for both general text and named entities compared to other open-source and proprietary LLMs, including those with significantly larger parameter counts.

\begin{figure}[t]
    \centering
    \includegraphics[width=\columnwidth]{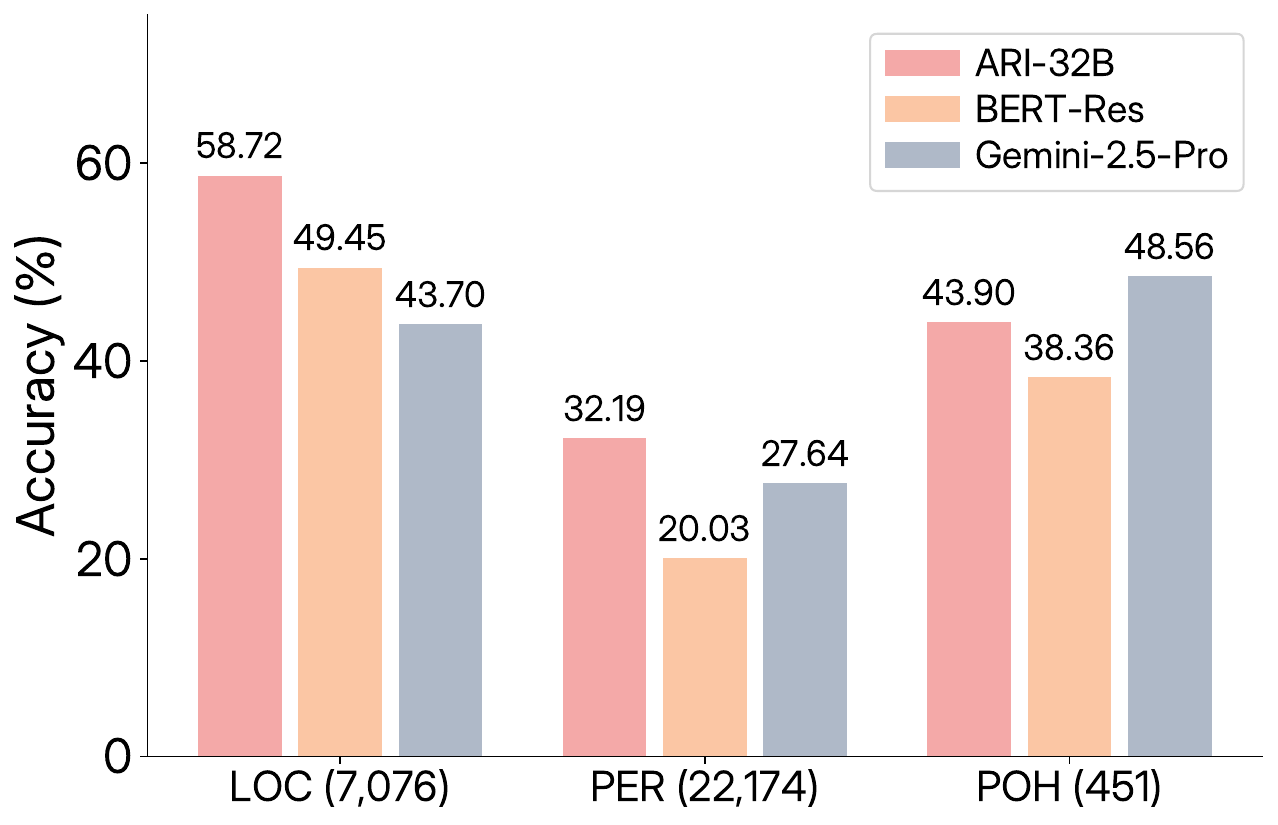}
    \caption{Top-1 accuracy comparison across different named entity categories on the $D_{\text{NE}}$.}
    \label{fig:ner_accuracy_by_category}
\end{figure}

We also conducted a fine-grained performance evaluation stratified by named entity types. First, we selected samples containing explicit named entity tags from the test dataset, $D_{\text{NE}}$. We then constructed a specialized evaluation dataset of 7,707 documents by artificially masking only the target named entity tokens. For named entity classification, we employed Gemini-2.5-Pro. The entities were categorized into Person (PER), Location (LOC), and Publication of History (POH), comprising 22,174, 7,076, and 451, respectively. Finally, we measured the Top-1 accuracy for each category to compare the restoration performance across models; the results are presented in Fig.~\ref{fig:ner_accuracy_by_category}.

First, ARI-32B demonstrated superior performance in named entity restoration, exceeding both BERT-Res and Gemini-2.5-Pro in the LOC and PER categories. Conversely, Gemini-2.5-Pro achieved the highest performance in the POH category, where such instances are less prevalent. Our qualitative analysis revealed that the POH category contains a significant number of Confucian classics, which originated in China and are widely shared across East Asia. This suggests that Gemini-2.5-Pro benefits from extensive knowledge of Chinese classics acquired through multilingual pre-training, unlike models specialized solely for the Joseon dynasty domain, resulting in relatively higher performance in this specific category.

\begin{table}[t]
    \centering
    \setlength{\tabcolsep}{4.5pt}
    \begin{tabular}{lccc}
        \toprule
          & ARI-32B & BERT-Res & Gemini \\
         \midrule
         Accuracy@1 & \textbf{0.383} & 0.207 & 0.273 \\
         Accuracy@10 & \textbf{0.583} & 0.403 & - \\
         nDCG@10 & \textbf{0.477} & 0.296 & - \\
         Win Ratio (\%) & \textbf{46.0} & 24.0 & 30.0 \\
         \bottomrule
    \end{tabular}
    \caption{Restoration performance of ARI-32B, BERT-Res, and Gemini-2.5-Pro as evaluated by human experts.}
    \label{tab:performance_expert}
\end{table}

\begin{table*}[t]
\setlength\doublerulesep{0.1cm}
  \setlength{\tabcolsep}{4.8pt}
  \centering
  {\fontsize{10.3pt}{12.5pt}\selectfont
  \begin{tabularx}{\textwidth}{c|X}
  \Xhline{2\arrayrulewidth}
  Damaged Text & 吏曹判書\textbf{\scalebox{0.86}[1.0]{[D1][D2][D3]}}初度呈辭. 入啓. \textbf{\scalebox{0.86}[1.0]{[D4]}}由. \hspace{2em} {\fontsize{10pt}{12.5pt}\selectfont (from JRS, 12/6/1654, King Hyojong)} \\
  \hline
  BERT-Res & 吏曹判書\textbf{\textcolor{red}{李壽恒}}初度呈辭. 入啓. {\textcolor{blue}{給}}由. \\
  \hline
  Gemini-2.5-Pro & 吏曹判書\textbf{\textcolor{red}{沈之源}}初度呈辭. 入啓. {\textcolor{blue}{給}}由. \\
  \hline
  ARI-32B & 吏曹判書\textbf{\textcolor{blue}{李厚源}}初度呈辭. 入啓. {\textcolor{blue}{給}}由. \\
  \hline \hline
  \multirow{2}{*}{Damaged Text} & 上詣敬奉閣行禮, 仍御承文院, 誦\scalebox{0.86}[1.0]{<\textbf{[D1][D2]}> <下泉>}詩, 命三學士及諸忠\textbf{\scalebox{0.86}[1.0]{[D3][D4]}}孫錄用. \\ 
  & {\fontsize{10pt}{12.5pt}\selectfont(from AJD, 10/6/1770, King Yeongjo)}  \\
  \hline
  BERT-Res & 上詣敬奉閣行禮, 仍御承文院, 誦<\textbf{\textcolor{red}{詩鑑}}> <下泉>詩, 命三學士及諸忠\textbf{\textcolor{blue}{臣子}孫}錄用. \\
  \hline
  Gemini-2.5-Pro & 上詣敬奉閣行禮, 仍御承文院, 誦<\textbf{\textcolor{blue}{匪風}}> <下泉>詩, 命三學士及諸忠\textbf{\textcolor{blue}{臣}}\textbf{\textcolor{red}{之}孫}錄用. \\
  \hline
  ARI-32B & 上詣敬奉閣行禮, 仍御承文院, 誦<\textbf{\textcolor{blue}{匪風}}> <下泉>詩, 命三學士及諸忠\textbf{\textcolor{blue}{臣}}\textbf{\textcolor{red}{後}孫}錄用. \\
  \hline \hline
  \multirow{2}{*}{Damaged Text} & 全羅道監司書狀, 去九月間, \textbf{\scalebox{0.86}[1.0]{[D1][D2]}}池水變赤色十餘日, 魚蝦浮出盡死事. \\ 
  & {\fontsize{10pt}{12.5pt}\selectfont(from AJD, 17/11/1588, King Seonjo)} \\
  \hline
  BERT-Res & 全羅道監司書狀, 去九月間, \textbf{\textcolor{red}{大山}}池水變赤色十餘日, 魚蝦浮出盡死事. \\
  \hline
  Gemini-2.5-Pro & \selectfont 全羅道監司書狀, 去九月間, \textbf{\textcolor{red}{海南}}池水變赤色十餘日, 魚蝦浮出盡死事. \\
  \hline
  ARI-32B & 全羅道監司書狀, 去九月間, \textbf{\textcolor{blue}{光州}}池水變赤色十餘日, 魚蝦浮出盡死事. \\
  \Xhline{2\arrayrulewidth}
\end{tabularx}
}
\vspace{-0.2cm}
\caption{Comparison of restoration examples between our model and baseline models.}
\label{tab:restoration_examples}
\end{table*}

\subsection{Evaluation by Human Experts on Real-World Damaged Documents}

To evaluate the efficacy of our restoration model as a practical collaborative tool, we engaged three experts in Sinographic literature and Korean history. We randomly sampled 100 documents from $D_{\text{RD}}$, which consists of real-world damaged documents. Remarkably, when classifying named entities using Gemini-2.5-Pro based on prior methods, we observed a high density of 74.6\%.

For the evaluation, we presented restoration candidates generated by ARI-32B, BERT-Res, and Gemini-2.5-Pro\footnote{As Gemini-2.5-Pro operates as a black-box model without logit access or beam search, we evaluated only its Top-1 candidate and excluded ranking-dependent metrics.} to experts in a blinded manner. Unlike the black-box nature of Gemini, ARI-32B and BERT-Res provide access to logits, enabling the generation of Top-$K$ candidates as described in Appendix~\ref{Appendix:top-k_candidates}. Experts selected valid candidates based on contextual coherence, grammatical correctness, and factual validity. Quantitatively, we designated these expert selections as the ground truth to compute Top-1 and Top-10 accuracy, as well as nDCG@10. We also measured the win rate to determine which model served as the most effective assistive tool, considering both the accuracy and diversity of the candidates.

As demonstrated in Table~\ref{tab:performance_expert}, ARI-32B outperforms both Gemini-2.5-Pro and BERT-Res in terms of Top-1 and Top-10 accuracy. Regarding nDCG@10, the model consistently ranked expert-validated candidates higher than BERT-Res. These results indicate that our model restores damaged documents more accurately than the baseline models. Additionally, it achieved a higher win ratio, reflecting a strong preference for its utility as a collaborative restoration tool. This suggests that ARI-32B effectively places reliable candidates at the top ranks while ensuring diversity, significantly reducing the search cost for experts in identifying the correct characters.

\subsection{Qualitative Analysis of Restoration}

To evaluate qualitative restoration performance, Table~\ref{tab:restoration_examples} presents restoration results from ARI-32B, Gemini-2.5-Pro, and BERT-Res on representative sample documents.

The first example involves the restoration of person names and general text. Although all models correctly recovered the general text in [D4], only ARI-32B successfully restored the person names in [D1], [D2], and [D3]. The second example focuses on book titles and general text. Both Gemini-2.5-Pro and ARI-32B successfully restored the book titles in [D1] and [D2]. For the general text in [D3] and [D4], only BERT-Res produced exact matches. However, our analysis confirmed that the prediction generated by ARI-32B, although differing from the ground truth ``子孫'' by producing ``後孫'', was semantically equivalent. The last example concerns the red tide phenomenon~\cite{lee2017red_tide}. In this case, ARI-32B demonstrated robustness in handling proper nouns; specifically, it was the only model to correctly restore the region names in [D1] and [D2]. Overall, these qualitative results suggest that ARI-32B outperforms the other models, particularly in restoring proper nouns such as personal and geographic names, by effectively leveraging external knowledge.

\subsection{Impact of Temporal Domain Shift}

\begin{figure}[t]
    \centering
    \includegraphics[width=\columnwidth]{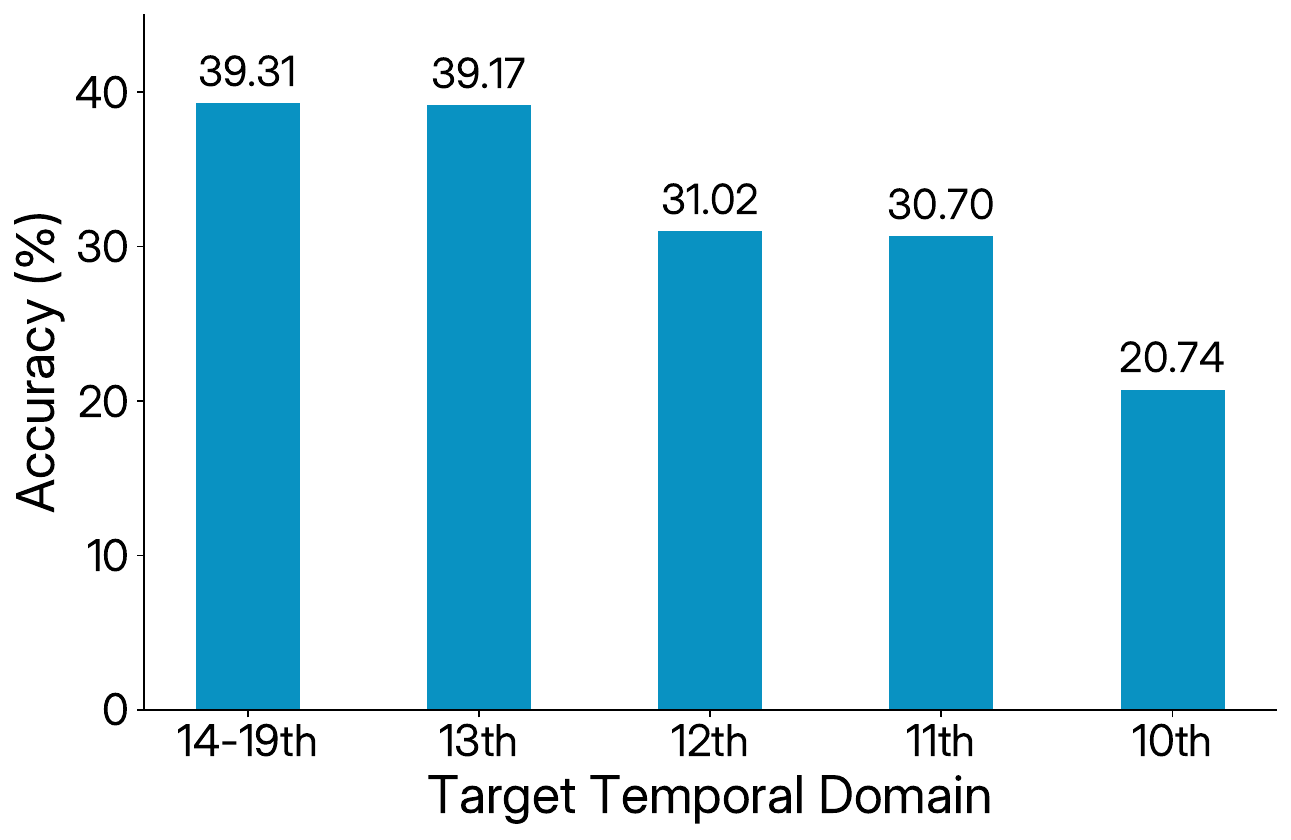}
    \caption{Restoration performance across unseen temporal domains, highlighting the impact of temporal shifts.}
    \label{fig:temporal_domain_shift}
\end{figure}

To investigate the impact of the temporal gap between the retrieved and target documents on ARI's performance, we additionally performed a restoration experiment on the \textit{Goryeosa}\footnote{\url{https://db.history.go.kr/goryeo}}, a historical chronicle of the Goryeo Dynasty, which predates our training dataset. Following the methodology of our survey, we constructed the test dataset from this corpus, employing the AJD and the JRS spanning the 14th to 19th centuries as reference corpora. As shown in Fig.~\ref{fig:temporal_domain_shift}, while ARI-32B remains robust under moderate temporal shifts, its performance declines as the temporal gap widens. This degradation is likely due to both diachronic linguistic changes and discrepancies in the period-specific external knowledge encapsulated within the reference corpora. Mitigating the impact of such domain mismatches remains an area for future research.
\section{Conclusion}

In this work, we introduced a RAG framework that enables LLMs to effectively utilize external knowledge for restoring damaged historical documents. Experimental results demonstrated consistent improvements in overall restoration performance, with especially notable gains in recovering named entities that depend on external knowledge. Furthermore, quantitative evaluation on a ground-truth test set and expert assessment on real-world damaged data showed that our model outperforms existing baselines and provides practical value for domain experts. Although developed for Korean historical archives, the proposed framework has broader potential as a general approach to low-resource ancient language restoration. By reducing the time and cost of restoring damaged historical documents, we expect this framework to help reveal previously obscured information and encourage the wider use of historical archives.

\section*{Limitations}

In this study, we constructed training, valid, and test datasets by aligning them with statistical characteristics, such as the frequency of damaged characters, observed in real-world documents. We additionally validated our model's performance through human-expert evaluation employing actual damaged documents. Despite these efforts, a potential discrepancy may persist between our constructed datasets and the intrinsic characteristics of real-world damaged documents.

Due to computational constraints, we limited the maximum input length of the model to 4,096 tokens. Consequently, approximately 1.7\% of data exceeding this length were filtered. In future work, we aim to extend the context window to accommodate longer documents and further improve performance. Beyond this, while our current approach leverages external knowledge and textual context, it relies solely on the text modality. Since the National Institute of Korean History provides original images for each document, incorporating these visual features into a multimodal restoration framework could significantly enhance performance.

\section*{Acknowledgments}
This work was supported by the National Research Foundation of Korea (NRF) grant funded by the Korea government (MSIT) (RS-2025-00555320) and the National Supercomputing Center with supercomputing resources including technical support (KSC-2024-CRE-0388).





\bibliography{reference}

\clearpage

\appendix

\section{Training Dataset Construction and Training Settings}
\label{sec:training_details}

In this study, we conducted a comparative experiment on training data mixing ratios to examine how each dataset affects model performance. The total size of the training set was fixed at 100K samples. Each configuration comprised a mixture of $D_{\text{NE}}$ and $D_{\text{Rand}}$, with the proportion of $D_{\text{NE}}$ varied across settings. As shown in Table~\ref{tab:train_mixing_ratio}, increasing the proportion of $D_{\text{NE}}$ consistently improved restoration performance on named entities, while random character restoration performance showed a downward trend. Considering this trade-off, the setting with 25\% $D_{\text{NE}}$ achieved the most balanced performance across the two metrics. Based on this result, we constructed the final training dataset using this ratio.

We trained ARI-32B on eight HGX H200 GPUs utilizing FSDP~\cite{zhao2023pytorch} and the Lion-8Bit optimizer~\cite{chen2023symbolic}. We employed a cosine learning rate schedule with a peak learning rate of $6 \times 10^{-6}$ and a warmup ratio of 0.05. Training was performed for two epochs with a global batch size of 64 and a maximum sequence length of 4,096 tokens, totaling 16.06 billion training tokens. ARI-8B was trained under the same settings, except that it used four HGX H200 GPUs and a global batch size of 48. For the BERT-Res baseline, we used ModernBERT-large and trained it on two RTX A6000 GPUs. We again used a cosine learning rate schedule, with a learning rate of $1 \times 10^{-4}$ and a warmup ratio of 0.05. The model was trained for 10 epochs with a global batch size of 1,024 and a maximum sequence length of 2,048 tokens, requiring approximately 12 hours.

\begin{table}[t]
    \centering
    \begin{tabular}{cccc}
        \toprule
          $D_{\text{Rand}}$ : $D_{\text{NE}}$ & Acc. on NE & Acc. on Rand \\
         \midrule
         100\% : 0\% & 42.75 & 75.67 \\
         75\% : 25\% & 45.41 & 76.13 \\
         50\% : 50\% & 45.06 & 75.12 \\
         25\% : 75\% & 45.89 & 74.58 \\
         0\% : 100\% & 45.25 & 69.68 \\
         \bottomrule
    \end{tabular}
    \caption{Performance comparison across different mixing ratios for constructing the training dataset.}
    \label{tab:train_mixing_ratio}
\end{table}

\section{Experiments}

\begin{figure}[t]
    \centering
    \includegraphics[width=\columnwidth]
    {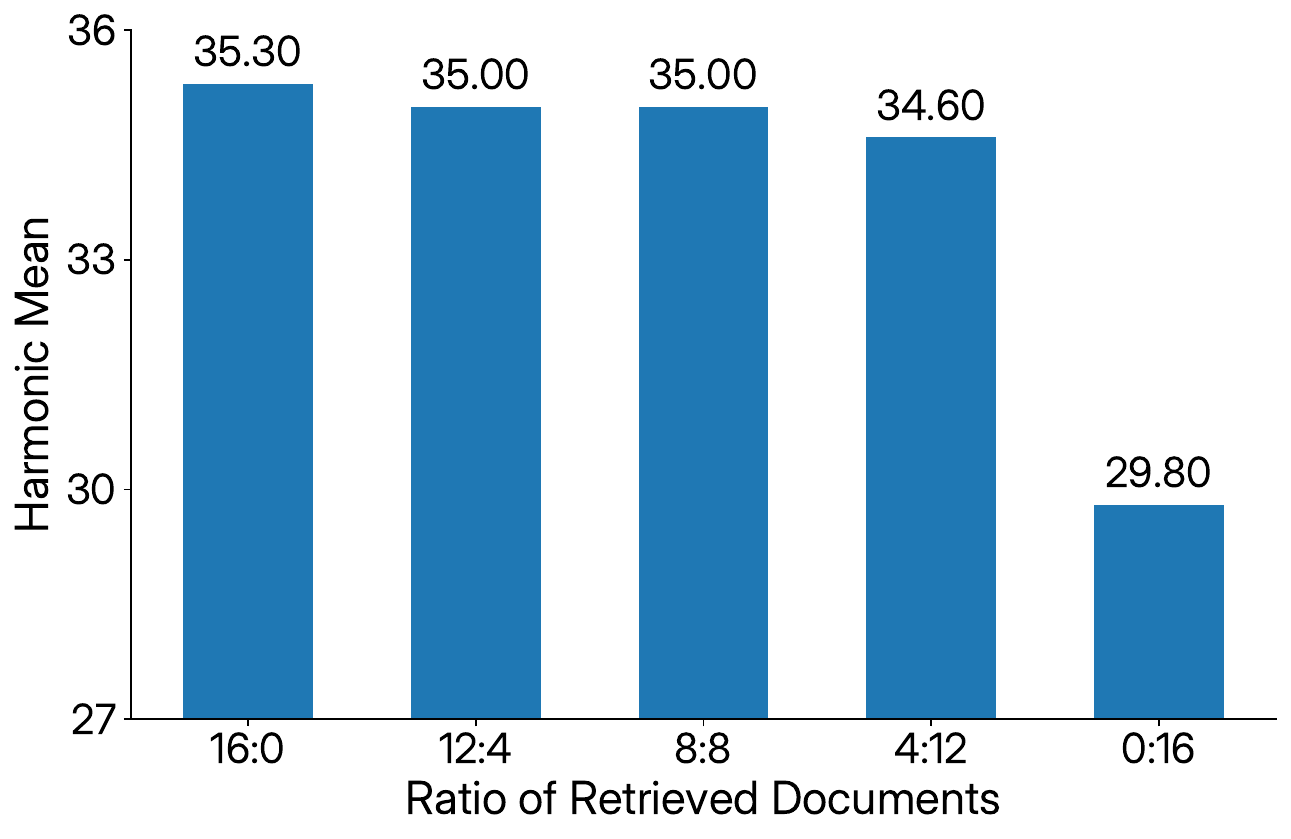}
    \caption{Restoration performance across different BM25-to-embedding retrieval ratios. The y-axis represents the harmonic mean of the scores on $D_{\text{NE}}$ and $D_{\text{Rand}}$ for each configuration.}
    \label{fig:shot_ratio_bar}
\end{figure}

\subsection{Analysis of Retrieval Strategies}
\label{sec:retrieval_analysis}
\begin{figure}[t]
    \centering
    \includegraphics[width=\columnwidth]{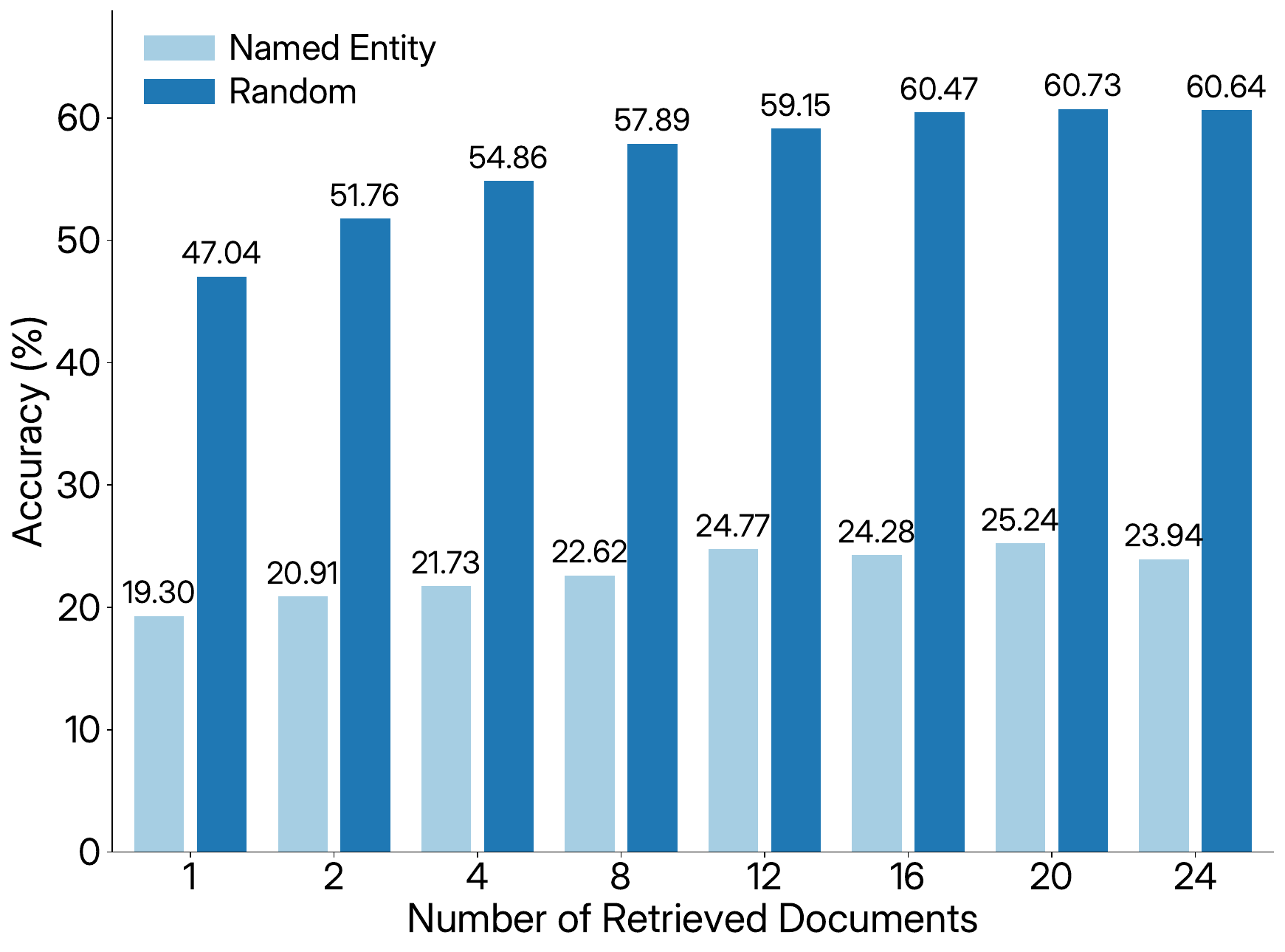}
    \caption{Performance comparison across different numbers of retrieved documents.}
    \label{fig:accuracy_by_num_retrieved_document}
\end{figure}

We further evaluated a hybrid RAG-based approach to determine the optimal RAG strategy for Hanja document restoration. We varied the composition of the 16 retrieved documents between BM25 and Gemini-Embedding. As shown in Fig.~\ref{fig:shot_ratio_bar}, the BM25-only configuration (i.e., a 16:0 ratio) achieved the best performance, yielding a harmonic mean of 35.3\%. This suggests that, given the character-level nature of the restoration task, lexical similarity plays a more important role than semantic similarity. Furthermore, we investigated a hybrid RAG strategy that additionally incorporated re-ranking using Qwen3-Reranker-8B~\cite{zhang2025qwen3}. This configuration achieved a $D_{\text{NE}}$ score of 37.70\%, which was lower than the 38.58\% obtained by the original BM25-based RAG pipeline. This result indicates that re-rankers pretrained on modern corpora have limitations in reordering retrieved Hanja documents for restoration tasks. Accordingly, we adopted a BM25-only strategy, prioritizing both restoration quality and architectural simplicity.

After determining the retrieval strategy, we further investigated the effect of varying the number of retrieved documents. Fig.~\ref{fig:accuracy_by_num_retrieved_document} illustrates the performance trends with respect to the number of retrieved documents. We find that, although performance generally improves on both $D_{\text{NE}}$ and $D_{\text{Rand}}$ as more documents are retrieved, it shows a slight decline when the number reaches 24. To balance computational efficiency and performance stability, we configure the ARI models to use 20 retrieved documents as few-shot examples.

\subsection{Top-k Accuracy during Training}

\begin{figure}[t]
    \centering
    \begin{subfigure}{\columnwidth}
        \includegraphics[width=\linewidth]{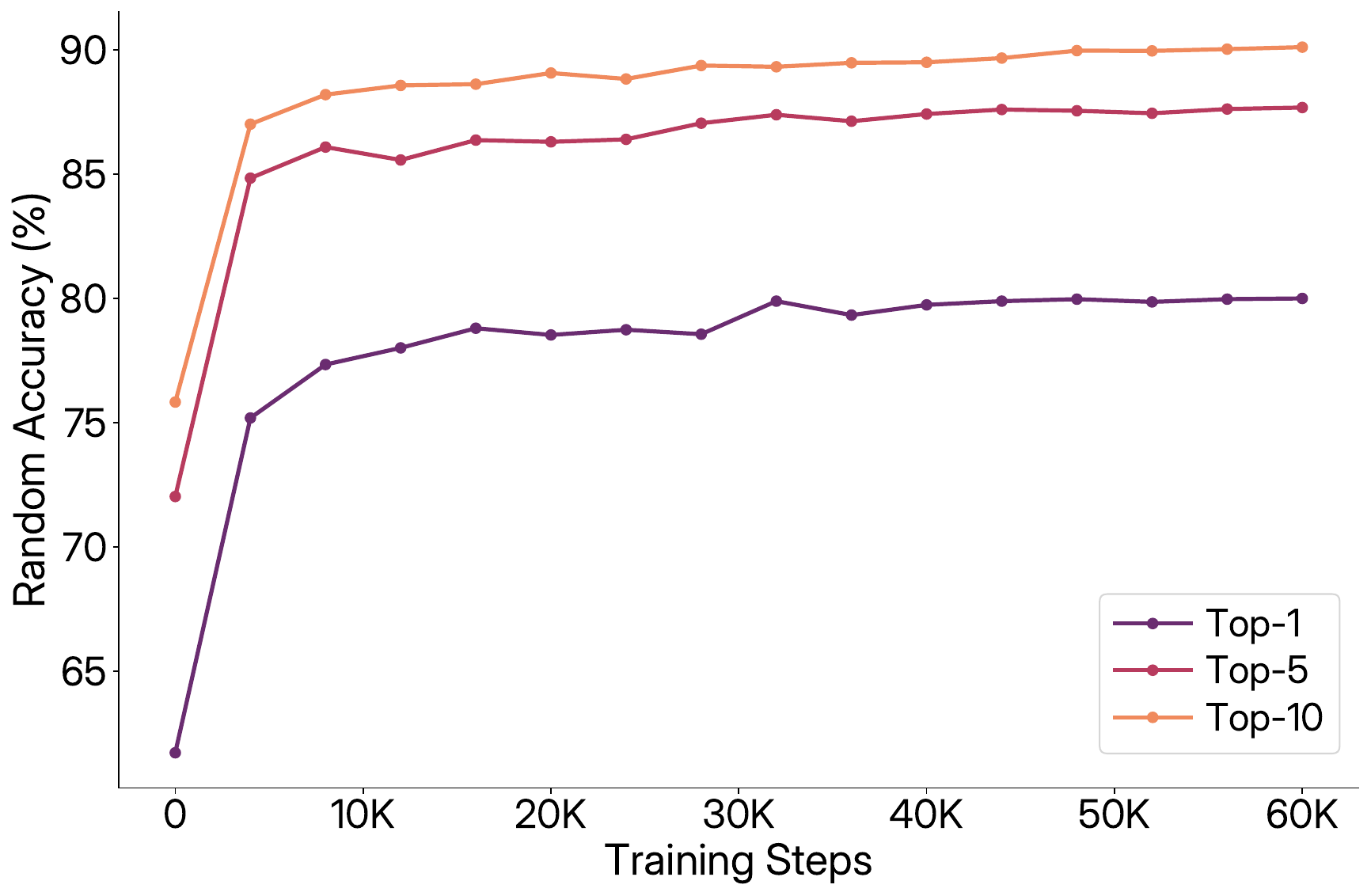}
    \end{subfigure}
    
    \vspace{1em}
    
    \begin{subfigure}{\columnwidth}
        \includegraphics[width=\linewidth]{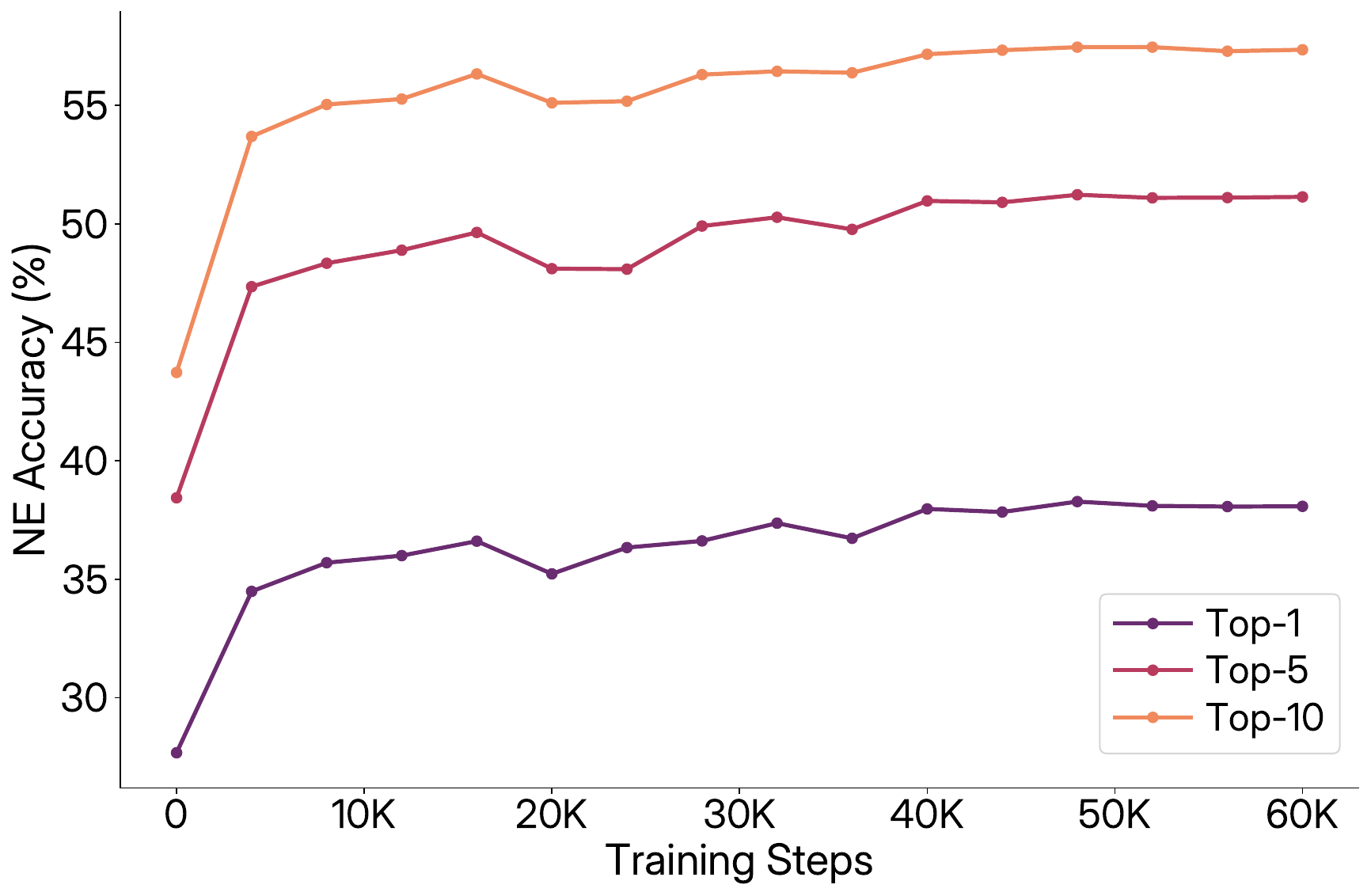}
    \end{subfigure}
        
    \caption{Fine-tuning accuracy of ARI-32B on the $D_{\text{Rand}}$ (top) and $D_{\text{NE}}$ (bottom) datasets.}
    \label{fig:topk_accuracy_by_checkpoint}
\end{figure}

Fig.~\ref{fig:topk_accuracy_by_checkpoint} illustrates the Top-1, Top-5, and Top-10 accuracies of ARI-32B measured on $D_{\text{Rand}}$ and $D_{\text{NE}}$, respectively, at each training step. These results demonstrate that the model's restoration performance improves as training progresses.

\section{Byte-level Constrained Decoding for Top-K Candidate Selection}
\label{Appendix:top-k_candidates}

\begin{figure}[t]
    \centering
    \includegraphics[width=\columnwidth]{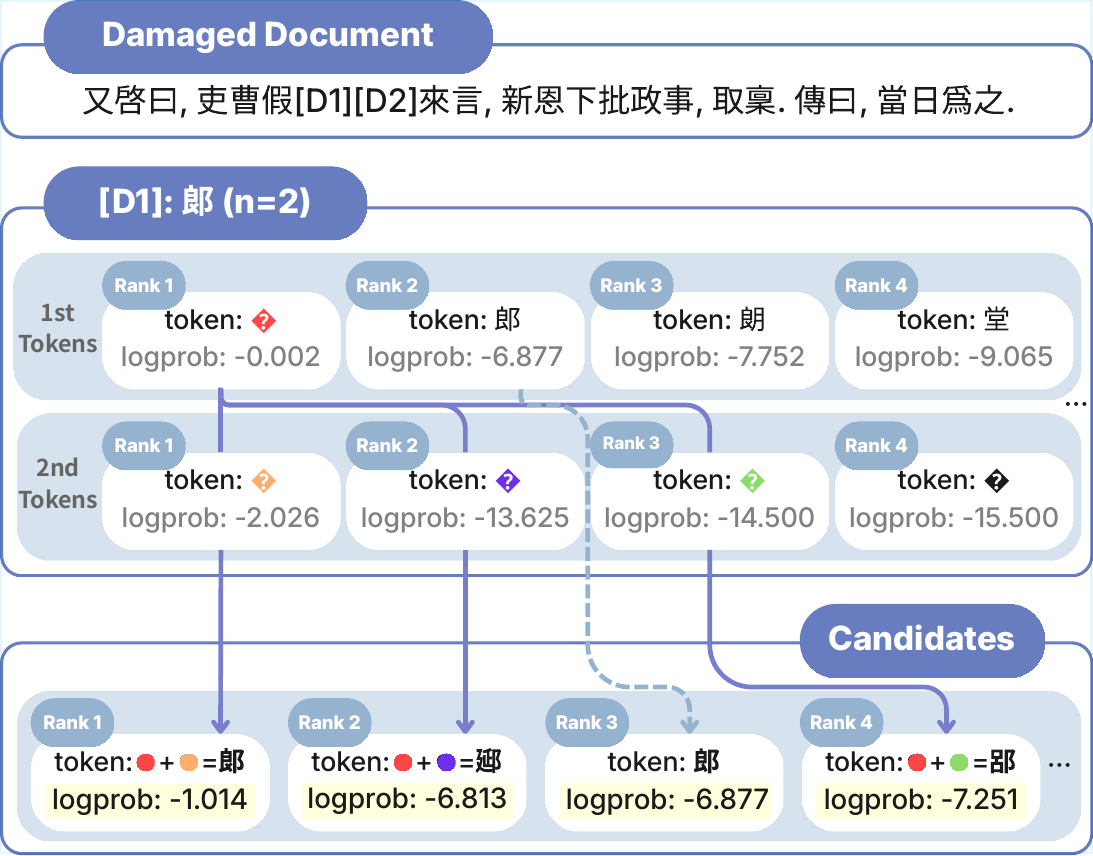}
    \caption{Illustration of Top-$K$ candidate selection via byte-level constrained decoding.}
    \label{fig:topk_candidates}
\end{figure}

With a Byte-level BPE tokenizer~\cite{wang2020neural}, frequent Chinese characters are typically mapped to single tokens. In contrast, Hanja characters, often absent from the tokenizer's vocabulary, are decomposed into sequences of multiple byte tokens (typically 2--3).

To address this fragmentation during the restoration of the $i$-th damaged character (illustrated in Fig.~\ref{fig:topk_candidates}), we employ a constrained decoding approach. Once an initial byte token is generated, the model continues generating tokens until a valid character sequence is formed. Candidate scores are then calculated by averaging the log probabilities of their constituent byte tokens, and the top-$K$ options are presented to the user. By default, the generation of the subsequent $(i+1)$-th character is conditioned on the top-1 prediction. However, if the user manually selects an alternative candidate (e.g., the $c$-th option), the context is updated to reflect this selection. We leverage the vLLM~\cite{kwon2023efficient} framework to efficiently extract these top-$K$ candidates.

\section{Expert Evaluation}

\begin{figure*}[t]
    \centering
    \includegraphics[width=0.8\linewidth]{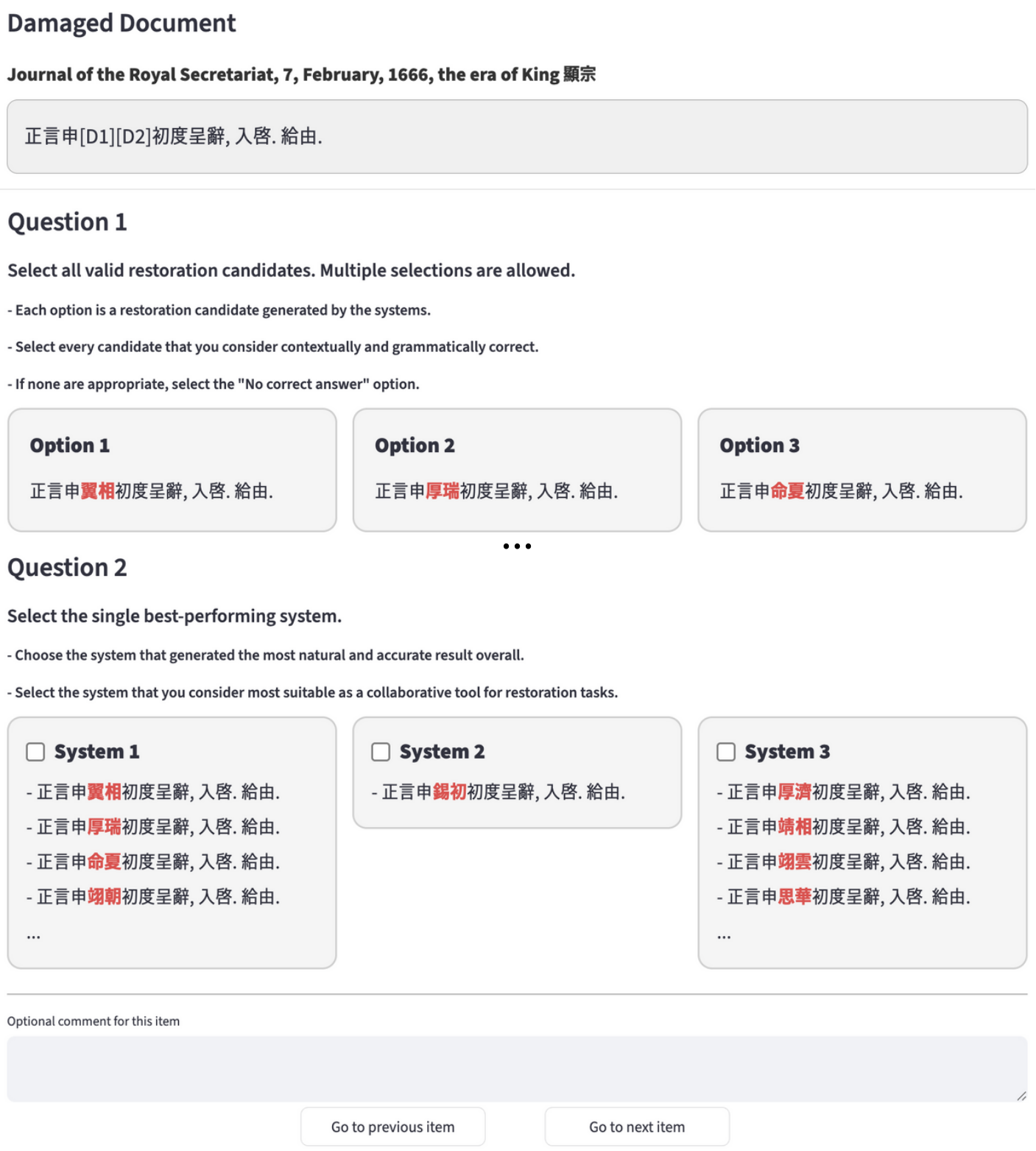}
    \caption{Web interface for expert evaluation.}
    \label{fig:expert_eval_ui}
\end{figure*}

To facilitate efficient expert evaluation, we implemented a web interface, as shown in Fig.~\ref{fig:expert_eval_ui}. The evaluation process consists of two questions for each real-world damaged document. First, evaluators are asked to select all valid answers from the candidates generated by different models (allowing for multiple selections), based on contextual coherence and factual validity. Subsequently, the second question investigates model preference; evaluators select the single most suitable model by comprehensively considering overall restoration quality and its utility as a collaboration tool.

\section{Prompt Format}
\label{sec:prompt}

Fig.~\ref{fig:final_prompt} presents the final prompt format, based on the experimental results in Section~\ref{external_knowledge}. The system prompt comprises four main components: Task, which defines the model's objective; Requirements, which outlines restoration guidelines tailored to the damaged document's context; Input \& Output, which dictates the interaction format; and Example Input \& Output, which provides few-shot examples. The user prompt reiterates basic instructions and supplies the target damaged document for restoration. Additionally, to enhance accuracy, we incorporate external knowledge comprising the document's corresponding date and 20 related documents retrieved from the training dataset.

\begin{figure*}[t]
    \centering
    \begin{tcolorbox}[
        title=\textbf{System Prompt},
        width=\textwidth,
        colframe=blue!30!gray,
        colback=blue!5!white,
        coltitle=white,
        boxrule=1pt,
        arc=1mm,
        boxsep=1pt,
        left=6pt, right=6pt, top=8pt, bottom=8pt,
        valign=top,
        toptitle=0.8mm,
        bottomtitle=0.8mm
    ]
    \small\ttfamily
    \raggedright
\vspace{2mm}

\#\#\# Task \\
You are an expert in restoring damaged Hanja characters. Restore each [Dn] with exactly the original Hanja character. Each [Dn] corresponds to exactly one Hanja character.
\\ 
\vspace{3mm}

\#\#\# Requirements \\
Base your restoration on the document’s overall context and meaning rather than treating each damaged token in isolation.
\\
\vspace{3mm}

\#\#\# Input \& Output \\
The input consists of the document itself, its metadata, and the related documents. \\
While the related documents are omitted in the shot for conciseness, they are always present in the actual dataset. \\
The output must follow the format: \{"[Dn]": "the restored Hanja character for [Dn]"\}
\\
\vspace{3mm}
\#\#\# Example Input \& Output \\

[Example 1] \\
**Input** \\ 
The document was written at date: 7, month: 6, year: 1771 - 英祖 era. \\
Input Document: "傳于[D1]興宗曰, 當自光明殿出, 承旨·侍衛, 來待于建禮門." \\
**Output** \\
\{"[D1]":"李"\} \\
\vspace{3mm}

[Example 2] \\
**Input** \\
The document was written at date: 26, month: 8, year: 1824 - 純祖 era. \\
Input Document: "[D1][D2]口傳政事, 副護軍單李紀淵." \\
**Output** \\
\{"[D1]":"兵","[D2]":"曹"\} \\
\vspace{3mm}

[Example 3] \\
**Input** \\
The document was written at date: 30, month: 6, year: 1686 - 肅宗 era. \\
Input Document: "府[D1]啓, 請[D2]禮·壽進·於[D3]·龍洞·明安公主房折受處[D4]査正事. 入啓." \\
**Output** \\
\{"[D1]": "前", "[D2]": "明", "[D3]": "義", "[D4]": "一"\} \\
\vspace{3mm}
...

\end{tcolorbox}

    \begin{tcolorbox}[
        title=\textbf{User Prompt},
        width=\textwidth,
        colframe=blue!30!gray,
        colback=blue!5!white,
        coltitle=white,
        boxrule=1pt,
        arc=1mm,
        boxsep=1pt,
        left=6pt, right=6pt, top=8pt, bottom=8pt,
        valign=top,
        toptitle=0.8mm,
        bottomtitle=0.8mm
    ]
    \small\ttfamily
    \raggedright
    
\vspace{2mm}
Use the following documents as references to accurately restore the input document. \\
\vspace{3mm}
Related Documents: \\
\vspace{3mm}
- 傳曰, 呈告工判·同敦寧許遞, 今日政差出.\\ 
- 傳曰, 在外敦寧都正·同敦寧, 許遞, 今日政差出.\\
- 傳曰, 呈告禮曹判書·同敦寧·兵曹參判·同成均許遞, 今日政差出.\\
- 傳于金敬均曰, 同敦寧許遞, 今日政差出.\\
- 傳于趙秉翊曰, 同敦寧·敦寧都正·同經筵許遞, 今日政差出. \\
\vspace{1mm}
...
\vspace{4mm}

The input document was written at date: 28, month: 7, year: 1887 - 高宗 era. \\ 
Input Document: 傳曰, 同敦寧·[D1][D2]都正竝許遞, 今日政差出.

\vspace{2mm}
    \end{tcolorbox}
    \caption{An example of the final prompt used for the restoration task.}
    \label{fig:final_prompt}
\end{figure*}

\end{document}